\documentclass[10pt,twocolumn,letterpaper]{article}

\usepackage{cvpr}
\usepackage{times}
\usepackage{epsfig}
\usepackage{bm}

\usepackage{floatrow}
\newfloatcommand{capbtabbox}{table}[][\FBwidth]

\usepackage{booktabs}       % professional-quality tables
\usepackage{nicefrac}       % compact symbols for 1/2, etc.
\usepackage{multirow}
\usepackage{amsmath}
\usepackage{amssymb}
\usepackage{array}

\cvprfinalcopy % *** Uncomment this line for the final submission

\def\cvprPaperID{5} % *** Enter the CVPR Paper ID here

\ifcvprfinal\fi

\title{Uncertainty Measures and Prediction Quality Rating for the Semantic Segmentation of Nested Multi Resolution Street Scene Images}

\author{Matthias Rottmann$^{1}$ \quad and \quad Marius Schubert$^{1}$ \\
\\
$^{1}$University of Wuppertal, School of Mathematics and Natural Sciences\\
\tt\small{\{\href{mailto:rottmann@uni-wuppertal.de}{rottmann},\href{mailto:marius.schubert@uni-wuppertal.de}{marius.schubert}\}@uni-wuppertal.de}
}

\usepackage{graphicx}
\usepackage{xcolor}

\usepackage[pagebackref=true,breaklinks=true,colorlinks,bookmarks=false]{hyperref}

\usepackage[nameinlink]{cleveref}
\DeclareMathOperator*{\argmax}{arg\, max}

\newcommand{\commentPS}[1]{#1}

\newcommand{\cropdist}{c_l}
\newcommand{\IoU}{\mathit{IoU}}
\newcommand{\mIoU}{\mathit{mIoU}}
\newcommand{\sIoU}{\mathit{IoU}_{\mathrm{adj}}}
\newcommand{\mean}{\mu}
\newcommand{\var}{v}

\usepackage{enumerate}

\newcommand{\intr}{{in}}
\newcommand{\bdr}{{bd}}

\begin{document}
\maketitle
\thispagestyle{empty}

\begin{abstract}
In the semantic segmentation of street scenes the reliability of the prediction and therefore uncertainty measures are of highest interest. We present a method that generates for each input image a hierarchy of nested crops around the image center and presents these, all re-scaled to the same size, to a neural network for semantic segmentation. The resulting softmax outputs are then post processed such that we can investigate mean and variance over all image crops as well as mean and variance of uncertainty heat maps obtained from pixel-wise uncertainty measures, like the entropy, applied to each crop's softmax output.
In our tests, we use the publicly available DeepLabv3+ MobilenetV2 network (trained on the Cityscapes dataset) and demonstrate that the incorporation of crops improves the quality of the prediction and that we obtain more reliable uncertainty measures. These are then aggregated over predicted segments for either classifying between $\IoU=0$ and $\IoU > 0$ (\emph{meta classification}) or predicting the $\IoU$ via linear regression (\emph{meta regression}). The latter yields reliable performance estimates for segmentation networks, in particular useful in the absence of ground truth. For the task of meta classification we obtain a classification accuracy of $81.93\%$ and an AUROC of $89.89\%$. For meta regression we obtain an $R^2$ value of $84.77\%$. These results yield significant improvements compared to other approaches.
\end{abstract}

\section{Introduction} \label{sec:intro}

\begin{figure*}[t]
    %\centering
    \includegraphics[width=\textwidth]{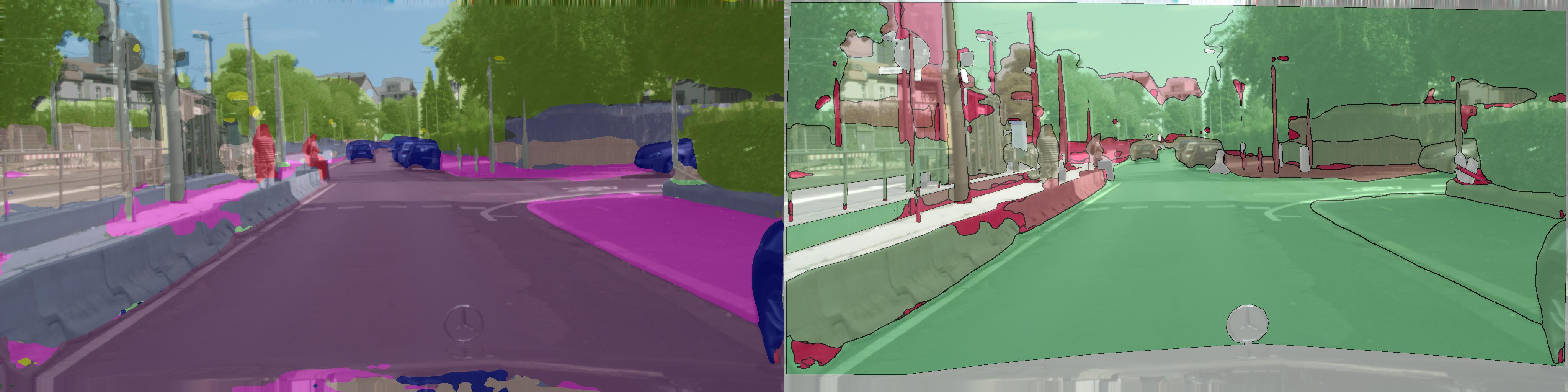}
    \caption{(Left): segmentation predicted by a neural network, (right): a visualization of the $\IoU$ which can only be computed in the presence of ground truth. Green color corresponds to high $\IoU$ values and red color to low ones, for the white regions there is no ground truth available. These regions are excluded from statistical evaluations.}
    \label{fig:iou_vis1}
\end{figure*}

In recent years, deep learning has outperformed other classes of predictive models in many applications. In some of these, e.g.\ autonomous driving or diagnostic medicine, the reliability of a prediction is of highest interest. In classification tasks, thresholding on the highest softmax probability or thresholding on the entropy of the classification distributions (softmax output) are commonly used approaches to detect false predictions of neural networks, see e.g.~\cite{DBLP:journals/corr/HendrycksG16c,DBLP:journals/corr/LiangLS17}. Metrics like classification entropy or the highest softmax probability are also combined with model uncertainty (Monte-Carlo (MC) dropout inference) or input uncertainty, cf.~\cite{Gal:2016:DBA:3045390.3045502} and \cite{DBLP:journals/corr/LiangLS17}, respectively. See \cite{oberdiek2018} for further uncertainty metrics. These approaches have proven to be practically efficient for detecting uncertainty and some of them have also been transferred to semantic segmentation tasks. The work presented in \cite{DBLP:journals/corr/KendallBC15} makes use of MC dropout to model the uncertainty of segmentation networks and also shows performance improvements in terms of segmentation accuracy. This approach was used in other works to model the uncertainty and filter out predictions with low reliability, cf.~e.g.~\cite{Kampffmeyer2016SemanticSO,DBLP:journals/corr/abs-1807-10584}. In \cite{huang2018efficient} this line of research was further developed to detect spacial and temporal uncertainty in the semantic segmentation of videos. In \cite{rottmann18} the concept of \emph{meta classification} in semantic segmentation, the task of predicting whether a predicted segment intersects with the ground truth or not, was introduced. This can be formulated as the task of classifying between $\IoU=0$ and $\IoU>0$ for every predicted segment (the $\IoU$ is also known as Jaccard index \cite{Jaccard12similarityCoefficient}). Furthermore a framework for the prediction of the $\IoU$ via linear regression (\emph{meta regression}) was proposed. The prediction of the $\IoU$ can be seen as a performance estimate which, after training a model, can be computed in the absence of ground truth. Both predictors use segment-wise metrics extracted from the segmentation network's softmax output as its input. A visualization of a segment-wise $\IoU$ rating is given in \cref{fig:iou_vis1}. Apart from the discussed uncertainty related methods, there are also works based on input image statistics. For instance, in \cite{rejectFP} a method for the rejection of false positive predictions is introduced. Performance measures for the segmentation of videos, also based on image statistics and boundary shapes, is introduced in \cite{perf_measure_video}.

In this work we elaborate on the uncertainty based approach from \cite{rottmann18} which is a method that consists of three simple steps. First, the segmentation network's softmax output is used to generate uncertainty heat maps, e.g.\ the pixel-wise entropy (cf.~\cref{fig:heatmaps}). In the second step, these uncertainty heat maps are then aggregated over the predicted segments and combined with other quantities derived from the predicted segments, e.g.\ the number of pixels per segment. From this we obtain segment-wise metrics. In the third step, these metrics are inputs for either a meta classification (between $\IoU=0$ and $\IoU>0$) or a meta regression for predicting the $\IoU$. In this paper, we perform the same prediction tasks, however we improve the method in all of its three steps.

\begin{figure}[t]
    \centering
    \includegraphics[width=\textwidth]{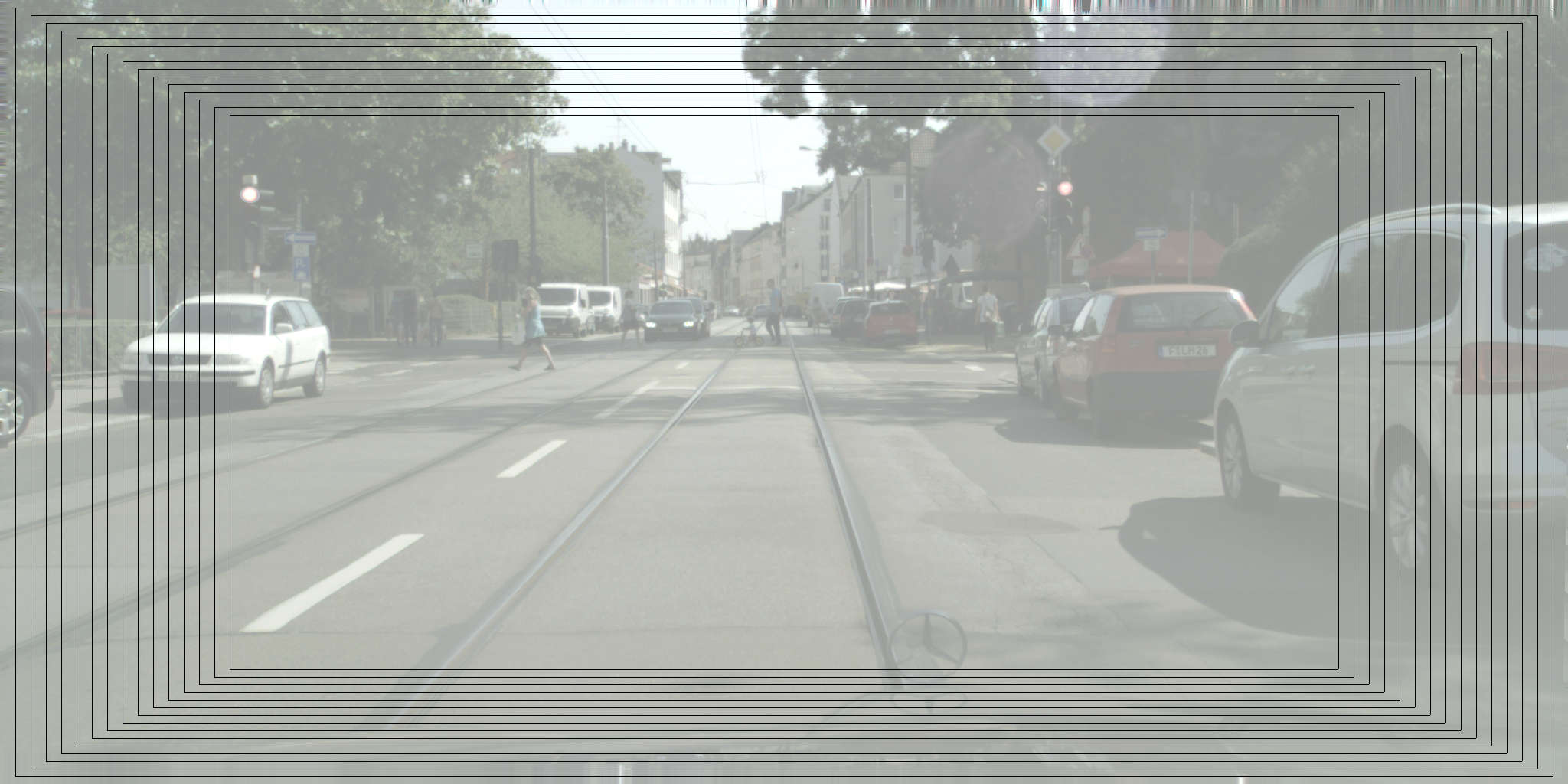}
    \caption{Visualization of a nested image cropping where all crops have the image center as focal point. This image is part of the Cityscapes dataset and has a resolution of $2048 \times 1024$ pixels. The original image is complemented with $15$ crops where each crop removes $c_l=10$ rows from the top and the bottom as well as the $20$ left-most and right-most columns of the previous crop.}
    \label{fig:nested_crops}
\end{figure}

In many scenarios, the camera system in use provides images with very high-resolution which are coarsened before presenting them to the segmentation network. Thus we loose information, especially for objects further away from the camera. Therefore we propose a method that constructs a hierarchy of nested image crops where all images have a common center point, see~\cref{fig:nested_crops}. All crops are then resized to the input size expected by the segmentation network such that we obtain an equally sized batch of input images. This can be processed by the neural network in a data parallel batch mode. Most neural network libraries, like e.g.~Tensorflow \cite{tensorflow2015-whitepaper}, are well vectorized over the input batch. Thus the increase in execution time should be below linear. The outputs of the segmentation network are then scaled back to its original size. In addition, we add kernel functions to let the crops smoothly fade into the combination of all larger crops, that have been merged with their predecessors recursively in the same way. We do this in order to avoid boundary effects. From this procedure we obtain a batch of probability distributions that are inputs to uncertainty measures, e.g.\ the entropy, probability margin and variation ratio. These are applied pixel-wise and yield heat maps for each probability distribution. A mean and a variance over all image crop heat maps give us additional uncertainty information compared to the uncertainty information used in \cite{rottmann18}.

Furthermore we elaborate on the approach from \cite{rottmann18} by introducing additional metrics that are derived from each segment's uncertainty and geometry information. In summary we end up with $42$ metrics (plus $19$ predicted class probabilities averaged over the predicted segments) in contrast to the $15$ metrics (plus $19$ class probabilities) introduced in \cite{rottmann18}. In addition to that, we study the incorporation of neural networks in meta classification and regression.

In our tests, we employ the publicly available DeepLabv3+ MobilenetV2 network \cite{deeplab,mobilenet} that was trained on the Cityscapes dataset \cite{cityscapes}. We perform all tests on the Cityscapes validation set. We demonstrate that the mean probability distribution over all crops provides improved $\IoU$ values and that the additional uncertainty heat maps, respectively their mean and variance, yield improved uncertainty information which results in better inputs for meta classification and regression. For the task of meta classification we obtain a classification accuracy of $81.93\%$ and an AUROC of $89.89\%$. For meta regression we obtain an $R^2$ value of $84.77\%$.  We also show that these results yield significant improvements compared to baseline approaches and the results obtained by the predecessor method introduced in \cite{rottmann18}.

The remainder of this work is structured as follows: In \cref{sec:nested_crops} we introduce the construction of the nested image crops, the aggregation of their softmax outputs and the resulting uncertainty heat maps. This is followed by the construction of segment-wise metrics using uncertainty and geometry information in \cref{sec:metrics}. Afterwards we present numerical results. First, we study the segmentation performance for different numbers of image crops. Then, we study how useful our segment-wise metrics are for meta classification and regression. This also includes a variable/metric selection study. Afterwards, we compare the meta classification and regression performance of our approach with baseline approaches and previous ones. Lastly, we study the incorporation of neural networks in meta classification and regression.

\section{Nested Image Crops and Uncertainty Measures} \label{sec:nested_crops}

Let $x \in \mathbb{R}^{N_r \times N_c \times 3} $ denote an RGB input image. For a chosen crop distance of $\cropdist$ we define a restriction operator $R_i$ that removes the $i \cdot \cropdist$ top and bottom rows as well as the $2i \cdot \cropdist$ left and right most pixels from $x$, i.e.,
\begin{align}
  R_i x = \{ x_{p,q,\cdot} \, : \;  & i\, \cropdist \leq p < N_r - i \, \cropdist,  \nonumber \\
                                    & 2i \, \cropdist \leq p < N_c - 2i \, \cropdist \} \, . 
\end{align}
In order to re-scale a cropped image to a desired resolution, we define an interpolation operator $I_i^j$ which performs a bi-linear interpolation for $ R_i x \in \mathbb{R}^{(N_r-2i \cropdist) \times (N_c - 4i\cropdist) \times 3} $ such that 
\begin{align}
& I_i^j R_i x \in \mathbb{R}^{(N_r-2j \cropdist) \times (N_c - 4j\cropdist) \times 3}
\nonumber \\
\text{and} \quad & I_i^0 R_i x \in \mathbb{R}^{N_r \times N_c \times 3} \, . \label{eq:bilinterp}
\end{align}
A segmentation network with a softmax output layer can be seen as a statistical model that provides for each pixel $z$ of the image a probability distribution $f_{z}(y|x,w)$ on the $C$ class labels $y\in\mathcal{C}=\{y_1,\ldots,y_C\}$.
\begin{equation}
P_i = \left( f_z(y | I_i^0 R_i x, w ) \right)_{ z \in \{1 , \ldots,  N_r\} \times \{1 , \ldots, N_c\} }
\end{equation}
for $i = 0, \ldots, N_\mathit{crop}$. Note that, due to \cref{eq:bilinterp}, i.e., all inputs being equally shaped, the $P_i$'s can be computed in batches which allows for efficient parallelization. In order to combine the probabilities $P_i$ to a common probability distribution we reshape them to their original size via
\begin{equation}
  Q_i = I_0^i P_i \, .
\end{equation}
We could now stack $Q_i$, $i=1,\ldots,N_\mathit{crop}$, in a pyramid fashion, sum them up and normalize the results such that we get a new probability distribution. However, this distribution would suffer from artifacts on the boundary of each $Q_i$. To avoid this, we proceed as follows: Let $Z_i$ define a zero padding operator such that $Z_i Q_i \in \mathbb{R}^{N_r \times N_c \times C}$ and $Q_i$ is centered in $Z_i Q_i$ while all other entries are zero. In order to construct a smooth mean probability distribution, we introduce a kernel function $K_i$ that is zero where $Z_i Q_i$ is zero and equal to one where the next nested crop $Z_{i+1} Q_{i+1}$ is not equal to zero. In-between these two regions, $K_i$ interpolates linearly. We can now recursively define our set of probability distributions, that we will use for further investigation, by
\begin{equation}
A_0 = P_0 \quad \text{and} \quad A_i = K_i Z_i Q_i + ( 1 - K_i) A_{i-1}
\end{equation}
for $i=1,\ldots,N_\mathit{crop}$. Each of the probability distributions $A_i$ can be viewed as a smooth merge of the current crop and the combination of all previously merged crops, due to their recursive definition being merged smoothly as well.

\begin{figure*}[t]
    \centering
    \includegraphics[width=\textwidth]{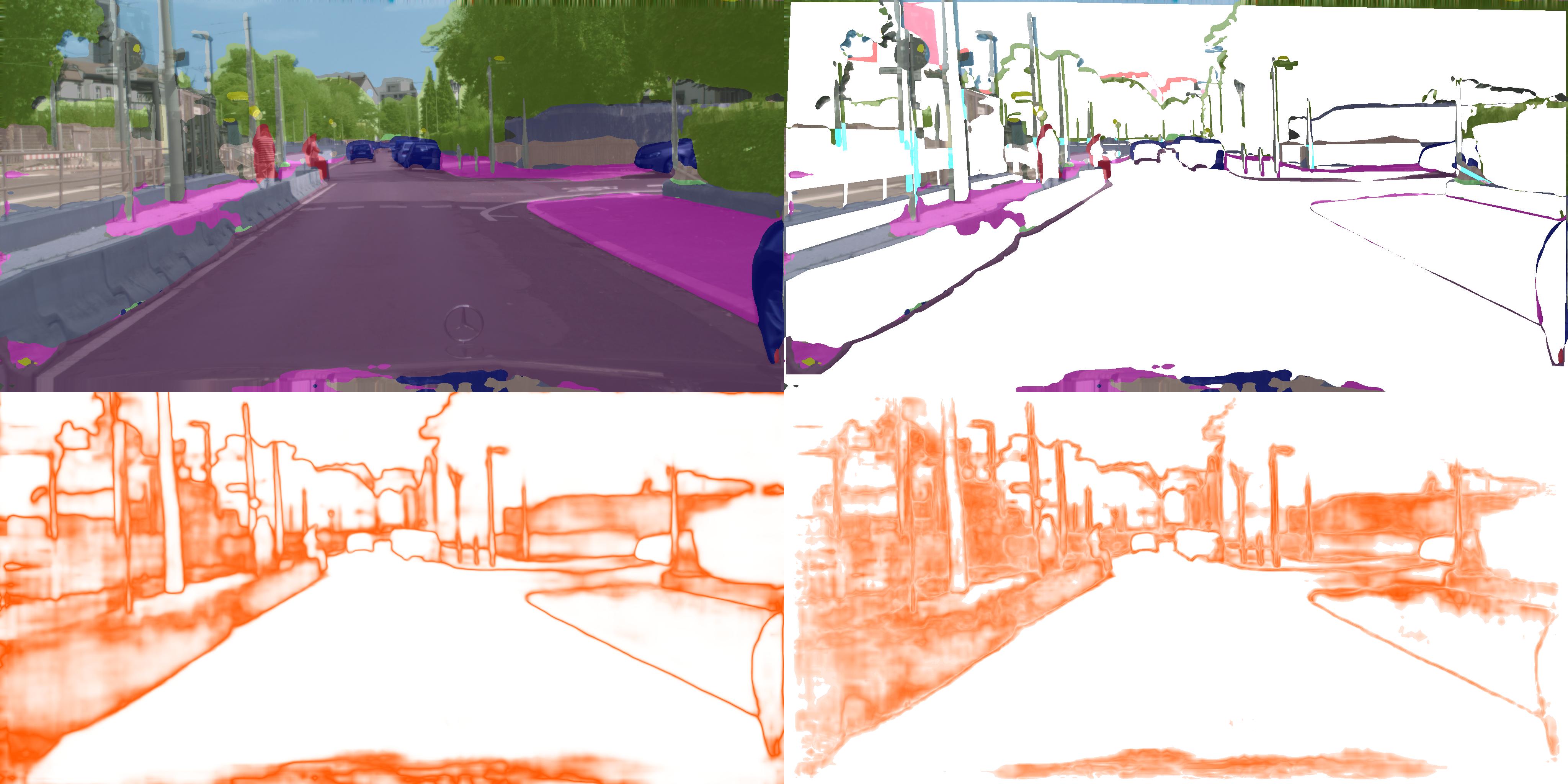}
    \caption{(Top left): segmentation $y_z(x)$ predicted by the neural network, (top right): predicted segmentation $y_z(x)$ where prediction and ground truth differ, note that the ego car is excluded from the ground truth, (bottom left): mean $\mean M_z$ of all probability margin heat maps, (bottom right): variance $\var M_z$ of probability margin heat maps.}
    \label{fig:heatmaps}
\end{figure*}

In the following we generate uncertainty heat maps for each $A_i$ by defining pixel-wise dispersion measures.
Let
\begin{equation}
\hat y_z(A_i)=\argmax_{y\in\mathcal{C}}A_{i,z,y}.
\end{equation}
denote the predicted class, for each pixel $z$ we define the \emph{entropy} (also known as \emph{Shannon information}~\cite{shannon1948}) $E_z$, the \emph{probability margin} $M_z$ and the \emph{variation ratio} $V_z$ by
\begin{align}
\label{eqa:entropy}
E_z(A_i)&=-\frac{1}{\log(C)}\sum_{y\in \mathcal{C}} A_{i,z,y} \log A_{i,z,y} \, , \\
\label{eqa:alt}
M_z(A_i) &= 1-A_{i,z,\hat y_z(A_i)} + \max_{y\in\mathcal{C}\setminus\{\hat y_z(A_i)\}} A_{i,z,y} \, , \\
\label{eqa:vr}
V_z(A_i) &= 1-A_{i,z,\hat y_z(A_i)} \, .
\end{align}
For each of these uncertainty measures $U_z \in \{ E_z, M_z, V_z \}$  we define a mean and a variance over the number of crops
\begin{align}
&\mean U_z = \frac{1}{N_\mathit{crop}} \sum_{i=0}^{N_\mathit{crop}} U_z(A_i) \nonumber \\
\text{and} \quad &
\var U_z = \mean ( U_z^2 ) - \mean( U_z )^2
\end{align}
Furthermore we also consider a symmetrized version of the Kullback-Leibler divergence of the mean probabilities $ A = \frac{1}{N_\mathit{crop}} \sum_{i=0}^{N_\mathit{crop}} A_i $ and the original probabilities $A_0$ without incorporation of additional crops, i.e.,
\begin{align}
K_z(A, A_0) = \frac{1}{2} \left( D_\mathit{KL}(A_z ||A_{0,z}) + D_\mathit{KL}(A_{0,z}|| A_z) \right)  \nonumber \\
= \frac{1}{2C} \sum_{y \in \mathcal{C}}  A_{z,y} \log ( \frac{A_{0,z,y}}{ A_{z,y}} ) + A_{0,z,y} \log ( \frac{ A_{z,y}}{A_{0,z,y}} ) \, .
\end{align}
A visualization of $\mean M_z$ and $\var M_z$ is given in \cref{fig:heatmaps}.
The heat maps $E_z, M_z, V_z$ and $K_z$ are subject to segment-wise investigation.

\section{Metrics Aggregated over Segments} \label{sec:metrics}

For a given image $x$ we define the set of connected components (segments) in the predicted segmentation $\hat{\mathcal S}_x=\{\hat y_z(A) | z\in x\}$ by $\hat{\mathcal K}_x$. Analogously we denote by ${\mathcal K}_x$ the set of connected components in the ground truth ${\mathcal S}_x$. For each $k\in \hat{\mathcal K}_x$, we define the following quantities:
\begin{itemize}
    \setlength\itemsep{-0.05em}
    \item the interior $k_\intr\subset k$ where a pixel $z$ is an element of $k_\intr$ if all eight neighbouring pixels are an element of $k$
    \item the boundary $k_\bdr=  k \setminus k_\intr$
    \item the intersection over union $\IoU$: let ${\mathcal K}_x|_k$ be the set of all $k'\in {\mathcal K}_x$ that have non-trivial intersection with $k$ and whose class label equals the predicted class for $k$, then
    $$
    \IoU(k) = \frac{|k \cap K'|}{|k \cup K'|}\,,\qquad K' = \bigcup_{k' \in {\mathcal K}_x|_k} k'
    $$
    \item adjusted $\sIoU$: let $Q = \{ q \in \hat{\mathcal{K}}_x: q \cap K' \neq \emptyset \} $, as in \cite{rottmann18} we use in our tests
    $$
    \sIoU(k) = \frac{|k \cap K'|}{|k \cup (K' \setminus Q)|}
    $$
    \item the pixel sizes $S=|k|$, $S_\intr=|k_\intr|$, $S_\bdr=|k_\bdr|$
    \item the mean dispersion $\bar D$, $\bar D_\intr$, $\bar D_\bdr$ defined as 
    $$\bar D_\sharp(k) = \frac{1}{S_\sharp} \sum_{z\in k_\sharp} D_z(x)\,,\qquad \sharp\in \{\_,in,bd\} $$
    where $D_z \in \{ K_z, \mean U_z, \var U_z \, : \, U_z = E_z, M_z, V_z \}$
    %\item the mean distances $\bar D$, $\bar D_\intr$, $\bar D_\bdr$ defined in analogy to the mean entropies
    \item the relative sizes $\tilde S = S/S_\bdr$, $\tilde S_\intr = S_\intr/S_\bdr$
    \item the relative mean dispersions $\tilde {\bar D} = \bar D \tilde S$, $\tilde {\bar D}_\intr = \bar D_\intr \tilde S_\intr$
    %, and relative mean distances $\tilde {\bar D} = \bar D \tilde S$, $\tilde {\bar D}_\intr = \bar D_\intr \tilde S_\intr$
    \item the geometric center $\bar k = (\bar k_1, \bar k_2) = \frac{1}{S} \sum_{z \in k} (z_1, z_2) $ where $z_1$ and $z_2$ are the vertical and horizontal coordinates of the pixel $z$ in $x$, respectively
    \item the mean class probabilities for each class $y \in \{1,\ldots,C\}$
    $$ P_y(k) = \frac{1}{S} \sum_{z \in k} A_{z,y} $$
    \item sets of metrics 
    $$\tau U = \{ \tau \bar U, \tau \bar U_\bdr, \tau \bar U_\intr, \tau \tilde{ \bar U }, \tau \tilde{ \bar U }_\intr \}$$
    for $\tau \in \{ \mean, \var \}$ and $U \in \{ V, M, E \}$ as well as
    $$P = \{ P_y \, : \, y=1,\ldots,C \}, \; \Sigma = \{ S, S_\intr, S_\bdr, \tilde S, \tilde S_\intr \} $$
\end{itemize}
\begin{table*}[bt]
\centering
\scalebox{0.78}{
\begin{tabular}{||l||c|c|c|c|c||c|c|c|c|c||}
\cline{1-11}
               & \multicolumn{5}{c||}{all $2048 \times 1024$ pixels}  &  \multicolumn{5}{c||}{$1024 \times 512$ center section} \\
\cline{1-11}
number of crops  & 1          & 2         & 4         & 8         & 16  &  1        &  2       &  4       &  8       &  16       \\
\cline{1-11}                                                                         
0: road           &  95.94\%  &  96.00\% &  96.04\% &  96.10\% &  \textbf{96.23\%}  &  95.00\%  &  95.05\% &  95.13\% &  95.25\% &  \textbf{95.52\%}  \\
1: sidewalk       &  71.83\%  &  72.08\% &  72.31\% &  72.63\% &  \textbf{73.26\%}  &  62.58\%  &  62.88\% &  63.27\% &  63.91\% &  \textbf{65.30\%}  \\
2: building       &  84.83\%  &  85.01\% &  85.15\% &  85.32\% &  \textbf{85.58\%}  &  76.79\%  &  77.07\% &  77.33\% &  77.70\% &  \textbf{78.43\%}  \\
3: wall           &  34.41\%  &  \textbf{34.48\%} &  34.40\% &  34.22\% &  33.92\%  &  32.55\%  &  32.97\% &  32.98\% &  \textbf{33.12\%} &  32.97\%  \\
4: fence          &  49.23\%  &  49.92\% &  49.96\% &  50.33\% &  \textbf{50.49\%}  &  41.07\%  &  41.48\% &  41.47\% &  42.24\% &  \textbf{42.90\%}  \\
5: pole           &  28.97\%  &  29.45\% &  29.89\% &  30.55\% &  \textbf{31.70\%}  &  22.06\%  &  22.50\% &  22.90\% &  23.72\% &  \textbf{25.59\%}  \\
6: traffic light  &  41.70\%  &  42.35\% &  42.72\% &  43.28\% &  \textbf{44.23\%}  &  26.40\%  &  27.56\% &  28.10\% &  29.00\% &  \textbf{30.85\%}  \\
7: traffic sign   &  50.59\%  &  50.94\% &  51.45\% &  52.08\% &  \textbf{53.27\%}  &  39.54\%  &  40.08\% &  40.95\% &  41.88\% &  \textbf{44.03\%}  \\
8: vegetation     &  84.43\%  &  84.58\% &  84.72\% &  84.90\% &  \textbf{85.23\%}  &  77.39\%  &  77.65\% &  77.92\% &  78.31\% &  \textbf{79.07\%}  \\
9: terrain        &  52.88\%  &  53.25\% &  53.43\% &  53.44\% &  \textbf{53.69\%}  &  43.88\%  &  44.46\% &  45.08\% &  45.49\% &  \textbf{46.25\%}  \\
10: sky            &  82.82\%  &  82.91\% &  82.98\% &  83.16\% &  \textbf{83.40\%}  &  64.91\%  &  65.07\% &  65.25\% &  65.83\% &  \textbf{67.20\%}  \\
11: person         &  63.40\%  &  63.85\% &  64.21\% &  64.93\% &  \textbf{66.11\%}  &  63.25\%  &  63.74\% &  64.20\% &  65.06\% &  \textbf{66.69\%}  \\
12: rider          &  43.63\%  &  43.90\% &  44.08\% &  44.50\% &  \textbf{45.41\%}  &  42.53\%  &  42.85\% &  43.15\% &  44.01\% &  \textbf{45.41\%}  \\
13: car            &  85.06\%  &  85.20\% &  85.40\% &  85.69\% &  \textbf{86.23\%}  &  79.38\%  &  79.58\% &  79.87\% &  80.37\% &  \textbf{81.37\%}  \\
14: truck          &  66.64\%  &  66.49\% &  66.41\% &  \textbf{65.82\%} &  64.16\%  &  66.97\%  &  67.54\% &  67.44\% &  \textbf{67.56\%} &  67.00\%  \\
15: bus            &  70.47\%  &  70.56\% &  \textbf{70.56\%} &  70.38\% &  70.22\%  &  70.95\%  &  71.17\% &  71.46\% &  71.60\% &  \textbf{71.85\%}  \\
16: train          &  58.44\%  &  59.63\% &  \textbf{59.92\%} &  58.87\% &  57.63\%  &  58.44\%  &  59.46\% &  60.51\% &  60.00\% &  \textbf{61.15\%}  \\
17: motorcycle     &  48.16\%  &  48.37\% &  48.63\% &  49.32\% &  \textbf{50.21\%}  &  45.21\%  &  45.49\% &  46.43\% &  47.28\% &  \textbf{48.57\%}  \\
18: bicycle        &  61.74\%  &  62.09\% &  62.44\% &  63.01\% &  \textbf{63.94\%}  &  55.22\%  &  55.73\% &  56.28\% &  57.09\% &  \textbf{58.65\%}  \\
\cline{1-11}
$\mIoU$        &  61.85\%  &  62.16\% &  62.35\% & 62.55\% & \textbf{62.89\%} &  56.01\%  &  56.44\% &  56.83\% &  57.34\% &  \textbf{58.36\%}  \\
\cline{1-11}
\end{tabular}
}
\caption{The (classical) $\IoU$ for each class over the whole dataset as well as the mean $\IoU$ ($\mIoU$) over all classes, both as a function of the number of crops. These numbers are computed once for the entire images of $2048\times1024$ pixels (left half) and once for the center section containing $1024 \times 512$ pixels (right-hand half). The best results for each class are highlighted.}
\label{tab:dataset_IoU}
\end{table*}
\begin{figure*}[t]
\begin{floatrow}
\capbtabbox{%
\scalebox{0.78}{{\setlength{\extrarowheight}{0.75pt}%
\begin{tabular}{|l|r||l|r||l|r||l|r|}
%\hline
%   & MN     &        & MN       &                &  MN      &   & MN       \\
\cline{1-8}
$\mean\bar E$               & \textbf{-0.71340} & $\var\bar E^*$               & -0.18668 & $\mean\bar M$               & \textbf{-0.84358} & $\var\bar M^*$               & -0.30971 \\
$\mean\bar E_\bdr$          & -0.43822 & $\var\bar E_\bdr^*$         & -0.14376 & $\mean\bar M_\bdr$          & -0.48518 & $\var\bar M_\bdr^*$          & +0.08374 \\
$\mean\bar E_\intr$         & \textbf{-0.71422} & $\var\bar E_\intr^*$         & -0.19332 & $\mean\bar M_\intr$         & \textbf{-0.83183} & $\var\bar M_\intr^*$         & -0.32423 \\
$\mean\tilde{\bar E}$       & +0.34611 & $\var\tilde{\bar E}^*$       & +0.33995 & $\mean\tilde{\bar M}$       & +0.30129 & $\var\tilde{\bar M}^*$       & +0.34914 \\
$\mean\tilde{\bar E}_\intr$ & +0.40510 & $\var\tilde{\bar E}_\intr^*$ & +0.37059 & $\mean\tilde{\bar M}_\intr$ & +0.34028 & $\var\tilde{\bar M}_\intr^*$ & +0.35836 \\
\cline{1-8}
$\mean\bar V^*$               & \textbf{-0.79546} & $\var\bar V^*$               & -0.36141 & $\bar K^*$                    & -0.33353 & $S$                        & +0.45958 \\
$\mean\bar V_\bdr^*$          & \textbf{-0.50218} & $\var\bar V_\bdr^*$          & -0.05362 & $\bar K_\bdr^*$               & -0.12983 & $S_\bdr$                   & \textbf{+0.60367} \\
$\mean\bar V_\intr^*$         & \textbf{-0.78578} & $\var\bar V_\intr^*$         & -0.36814 & $\bar K_\intr^*$              & -0.32906 & $S_\intr$                  & +0.45705 \\
$\mean\tilde{\bar V}^*$       & +0.25307 & $\var\tilde{\bar V}^*$       & +0.29991 & $\tilde{\bar K}^*$            & +0.17631 & $\tilde{S}$                & \textbf{+0.68636} \\
$\mean\tilde{\bar V}_\intr^*$ & +0.31223 & $\var\tilde{\bar V}_\intr^*$ & +0.32238 & $\tilde{\bar K}_\intr^*$      & +0.21686 & $\tilde{S}_\intr$          & \textbf{+0.68636} \\
\cline{1-8}
$\bar k_1^*$                  & -0.05955 & $\bar k_2^*$                 & +0.14190 & & & & \\
\cline{1-8}
\end{tabular}
}}
}{\caption{Pearson correlation coefficients for all constructed segment-wise metrics. All metrics marked with a ${}^*$ were not used in \cite{rottmann18}. All results with bsolute values greater than $0.5$ are highlighted.  \label{tab:corr_coeff}}%
}
\ffigbox[0.379\textwidth]{%
  \centerline{\includegraphics[width=.372\textwidth]{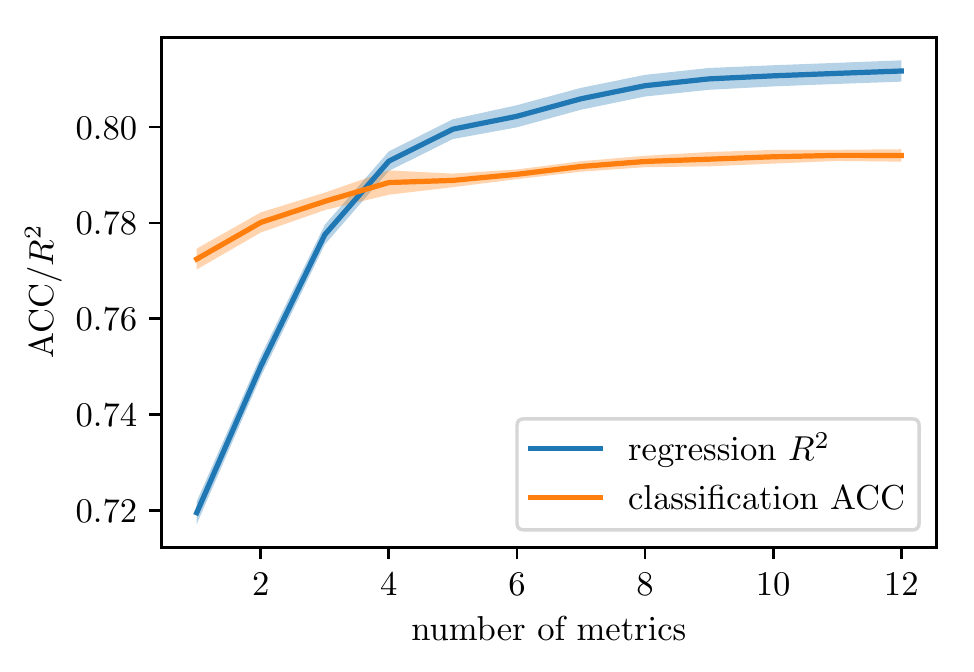}}
}{%
  \caption{Meta classification accuracy and meta regression $R^2$, both as a function of the number of metrics (sets stated in \cref{tab:greedy_var_select}). Results averaged over $10$ runs, the shaded regions depict the corresponding standard deviation. \label{fig:var_select}
  }
}
\end{floatrow}
\end{figure*}
Typically, \commentPS{$D_z$ is large for $z\in k_\bdr$}. This motivates the separate treatment of interior and boundary in all dispersion measures. Furthermore we observe that bad or wrong predictions often come with fractal segment shapes (which have a relatively large amount of boundary pixels, measurable by $\tilde S = S/S_\bdr$ and $\tilde S_\intr = S_\intr/S_\bdr$ ) and/or high dispersions $\bar D_\intr$ on the segment's interior. With the exception of $\IoU$ and $\sIoU$, all scalar quantities defined above can be computed without the knowledge of the ground truth. Our aim is to analyze to which extent they are suited for the tasks of meta classification and meta regression for the $\sIoU$.

\section{Numerical Experiments: Street Scenes} \label{sec:numexp}

\begin{table*}[t]
\centering
\scalebox{0.78}{{\setlength{\extrarowheight}{0.75pt}%
\begin{tabular}{||l||c|c||c|c||c|c||}
\cline{1-7}
\multicolumn{7}{||c||}{Meta Classification $IoU_{adj}=0,>0$} \\
\cline{1-7}
                             & \multicolumn{2}{c||}{entropy} & \multicolumn{2}{c||}{probability margin}                & class probabilities & \\
\cline{1-7}
                             & $\mean E \cup \var E$  & $\mean E$ & $\mean M \cup \var M$  & $\mean M$               & $P$ & \\
\cline{1-7}                                                                         
ACC              & $77.82\%(\pm0.26\%)$ & $77.06\%(\pm0.26\%)$          & $78.49\%(\pm0.28\%)$ & $76.99\%(\pm0.27\%)$  & $64.70\%(\pm0.36\%)$ & \\
AUROC            & $85.39\%(\pm0.21\%)$ & $84.66\%(\pm0.19\%)$          & $85.47\%(\pm0.21\%)$ & $85.06\%(\pm0.22\%)$  & $64.65\%(\pm0.34\%)$ & \\
\cline{1-7}                                                                     
\multicolumn{7}{||c||}{Meta Regression $IoU_{adj}$} \\                     
\cline{1-7}
$\sigma$         & $0.162(\pm0.001)$    & $0.163(\pm0.001)$             & $0.147(\pm0.001)$    & $0.150(\pm0.001)$     & $0.276(\pm0.001)$ & \\
$R^2$            & $73.59\%(\pm0.29\%)$ & $73.10\%(\pm0.28\%)$          & $78.27\%(\pm0.24\%)$ & $77.34\%(\pm0.23\%)$  & $22.92\%(\pm0.32\%)$ & \\
\cline{1-7}
\cline{1-7}
\multicolumn{7}{||c||}{Meta Classification $IoU_{adj}=0,>0$} \\
\cline{1-7}
                             & \multicolumn{2}{c||}{variation ratio} & \multicolumn{2}{c||}{segment sizes}            & \multicolumn{2}{c||}{all metrics}\\
\cline{1-7}
                             & $\mean V \cup \var V$  & $\mean V$ &    $\Sigma \cup \{ \bar k_1, \bar k_2 \}$ & $\Sigma$ &  with variances & without  \\
\cline{1-7}                                                                         
ACC              & $78.14\%(\pm0.25\%)$ & $76.96\%(\pm0.24\%)$          & $77.60\%(\pm0.17\%)$ & $77.25\%(\pm0.23\%)$  & $\boldsymbol{79.58\%(\pm0.15\%)}$ & $79.30\%(\pm0.11\%)$  \\
AUROC            & $85.41\%(\pm0.21\%)$ & $84.89\%(\pm0.21\%)$          & $84.94\%(\pm0.17\%)$ & $84.36\%(\pm0.25\%)$  & $\boldsymbol{87.38\%(\pm0.16\%)}$ & $87.08\%(\pm0.16\%)$  \\
\cline{1-7}                                                                     
\multicolumn{7}{||c||}{Meta Regression $IoU_{adj}$} \\                     
\cline{1-7}
$\sigma$         & $0.154(\pm0.001)$    & $0.156(\pm0.001)$             & $0.174(\pm0.001)$    & $0.179(\pm0.001)$     & $\boldsymbol{0.135(\pm0.001)}$    &  $0.136(\pm0.001)$    \\
$R^2$            & $76.12\%(\pm0.26\%)$ & $75.50\%(\pm0.26\%)$          & $69.41\%(\pm0.27\%)$ & $67.79\%(\pm0.27\%)$  & $\boldsymbol{81.71\%(\pm0.20\%)}$ &  $81.36\%(\pm0.19\%)$ \\
\cline{1-7}                                                                       
\end{tabular}   
}}
\caption{Comparison of sets of metrics. Each of the uncertainty heat map based set of metrics is used once including the variance metrics ($\mean U \cup \var U$ for $U=E,M,V$) and once without variance based metrics ($\mean U$ for $U=E,M,V$). We state results for the segments sizes $\Sigma$ including the geometric center $\bar k$ and without. The average predicted class probabilities $P$ are given by $19$ metrics, one for each class. All results are calculated on the metrics' validation set, the best results are highlighted.
%\commentMR{TODO: caption anpassen.}
}
\label{tab:subset_comparison}
\end{table*}

\begin{table*}[t]
\centering
\scalebox{0.70}{{\setlength{\extrarowheight}{0.75pt}
\begin{tabular}{||l||c|c|c|c|c|c|c|c|c|c|c|c|c||}
\cline{1-14}
number of metrics  & 1                & 2                      & 3              & 4                    & 5                    & 6              &  7                 &  8           &  9         &  10                  &  11                   & 12      & 61       \\
\cline{1-14}
classification accuracy (in \%)          & 0.7725                 & 0.7801               &  0.7845           & 0.7884         & 0.7889               & 0.7901         & 0.7918             & 0.7928       & 0.7933       & 0.7938           & 0.7941                 & 0.7944               & 0.7958 \\
added metric       & $\tilde{S}$  & $v{\hat M}_\intr$  & ${\bar k_2}$  & $P_{4}$  & $P_{5}$  & $v\tilde{\hat E}_\intr$  & $P_{14}$  & $P_{17}$  & $P_{15}$  & $P_{3}$  & $\mean{\bar M}_\bdr$  & $v\tilde{\hat E}$  & all \\
\cline{1-14}
regression $R^2$ (in \%)              & 0.7195           & 0.7501                 & 0.7776         & 0.7929               & 0.8000               & 0.8023         & 0.8059             & 0.8086       & 0.8101     & 0.8107            &  0.8112                 & 0.8117  & 0.8171 \\
added metric       & $\mean\bar M$  & $\tilde{S}$  & $\mean{\bar M}_\bdr$  & ${\bar k_2}$  & $\mean{\bar M}_\intr$  & $v{\hat M}_\bdr$ & $v\tilde{\hat E}$ & $P_{5}$  & $P_{11}$  & $P_{18}$   & $P_{0}$  & $\tilde{\bar K}_\intr$  & all  \\
\cline{1-14}
\end{tabular}
}}
\caption{Metric selection using a greedy method that adds in each step one metric that maximizes the meta classification/regression performance. The upper part of the table contains the sequence of metrics added corresponding to classification accuracy maximization, the lower one corresponding to $R^2$ maximization. All results are calculated on the metrics' validation set.}
\label{tab:greedy_var_select}
\end{table*}

In this section we investigate the properties of the nested crops and the metrics defined in the previous sections for the example of a semantic segmentation of street scenes. To this end, we consider the DeepLabv3+ network \cite{deeplab} with MobilenetV2 \cite{mobilenet} encoder for which we use a reference implementation in Tensorflow \cite{tensorflow2015-whitepaper} as well as weights pretrained on the Cityscapes dataset \cite{cityscapes} (available on GitHub). As parameters for the DeepLabv3+ framework we use an output stride of $16$, the input image is evaluated within the framework only on its original scale. These parameters result in a mean $\IoU$ of $61.85\%$ on the Cityscapes validation set, here mean refers to mean over all classes. We refer to \cite{deeplab} for a detailed explanation of the chosen parameters.

For our tests we produced $N_\mathit{crop}=15$ crops, i.e., we have $16$ nested images for each original image. The Cityscapes validation dataset contains $500$ images with a resolution of $2048\times1024$ pixels. Each crop is obtained from the previous one by removing the $20$ left-most and the $20$ right-most columns as well as the $10$ top and the $10$ bottom rows. In all tests we only consider segments with non-empty interior. For the combined prediction using all 16 crops, MobilenetV2 predicts $46896$ segments of which $38811$ have non-empty interior. From those segments with non-empty interior, $24354$ have $\sIoU > 0$. This gives a meta classification accuracy baseline of $62.75\%$ if we predict that each segment has $\sIoU > 0$. Note that, when only using the prediction of the original image, we obtain $53424$ components, $42261$ with non-empty interior of which $24590$ have $\sIoU > 0$ (resulting in $58.19\%$ meta classification baseline accuracy). Thus, meta classification results for different numbers of crops are not straight forward comparable. Hence, we focus on results for $16$ crops in the following studies.

All results, if not stated otherwise, were computed from $10$ repeated runs where training and validation sets (both of the same size) were re-sampled. We give mean results as well as corresponding standard deviations in brackets.

%\vspace{-2.5ex}
%\vspace{1ex} \noindent
%\textbf{Performance depending on the number of crops.}
\paragraph{Performance depending on the number of crops.}
\Cref{tab:dataset_IoU} contains the values for the classical $\IoU$ over the whole Cityscapes validation dataset for the different classes as a function of the number of crops (1,2,4,8,16), for the entire image ($2048\times1024$ pixels) as well as for the $1024\times512$ center pixels. In both cases the $\mIoU$ increases continuously when adding further crops. For the whole image the $\mIoU$ increases from $61.85\%$ to $62.89\%$ (i.e., by $1.04$ percentage points (pp)) and for the center section from $56.01\%$ to $58.36\%$ (by $2.35$ pp). This demonstrates that our crop based method indeed has the desired effect on smaller objects further away from ego car. For classes of particular interest, like person, rider and traffic sign, we observe improvements in the center section of $2.88$ (for rider) to $4.49$ pp (for traffic sign). We make these observations even though the original image is presented to the segmentation network at full resolution and the zoomed crops do not contain any additional information. In summary these results already justify the deployment of our approach which can be nicely parallelized over the data batch. In addition we obtain further uncertainty information which we investigate in the subsequent paragraphs.

\begin{figure*}[t]
\centerline{\includegraphics[width=1.0\textwidth]{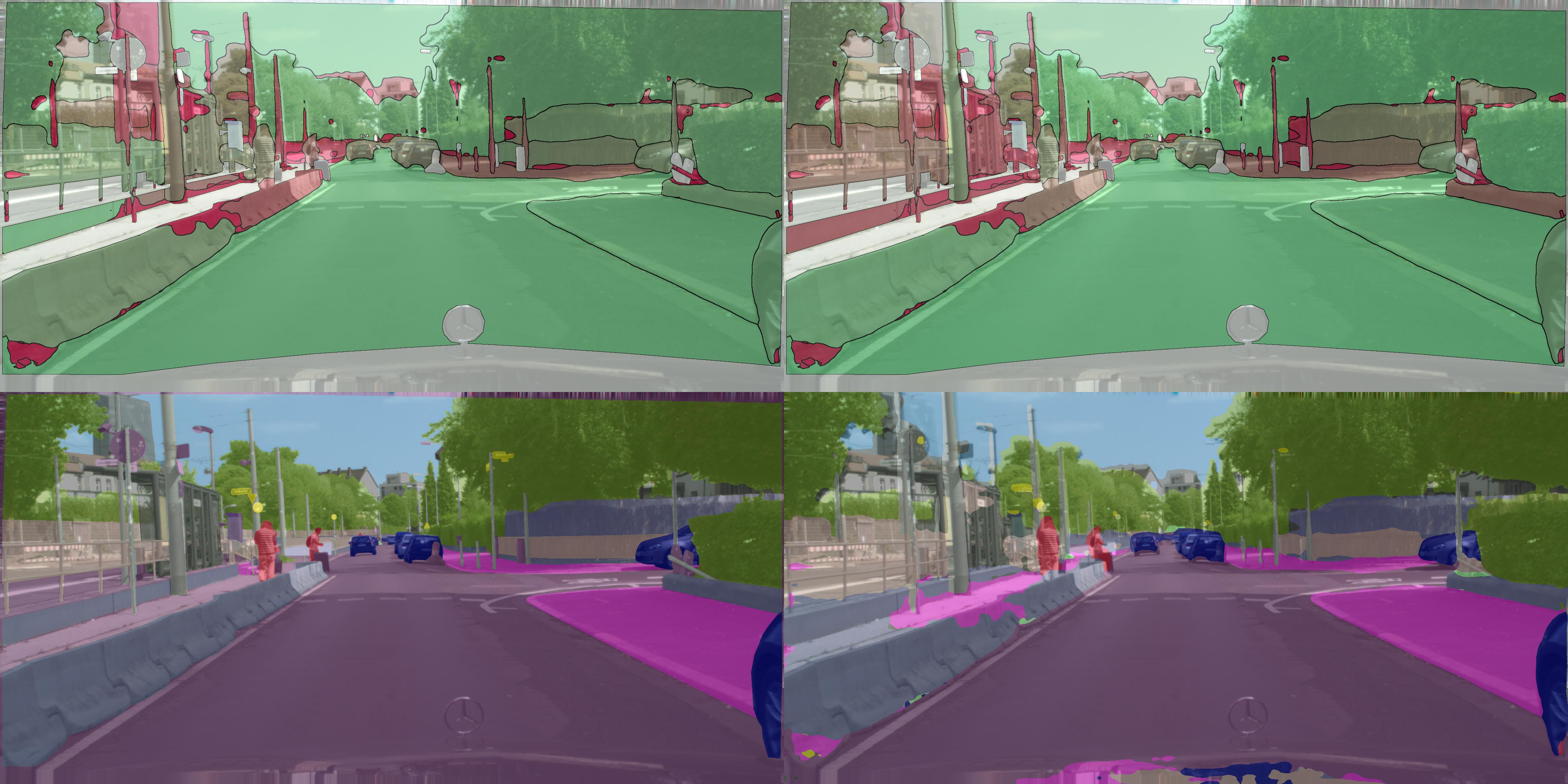}}
\caption{Prediction of the $\sIoU$ via linear regression. (bottom left): ground truth, (bottom right): predicted segments, (top left): true $\sIoU$ for the predicted segments and (top right): predicted $\sIoU$ for the predicted segments. In the top row, green color corresponds to high $\sIoU$ values and red color to low ones, for the white regions there is no ground truth available. These regions are excluded from the statistical evaluation.} \label{fig:illus1}
\end{figure*}

\begin{table*}[bt]
\centering
\scalebox{0.777}{{\setlength{\extrarowheight}{0.75pt}%
\begin{tabular}{||l||c|c||c|c||c|c||}
\cline{1-7}
\multicolumn{7}{||c||}{Meta Classification $IoU_{adj}=0,>0$} \\
\cline{1-7}
                  & \multicolumn{2}{c||}{all metrics}               & \multicolumn{2}{c||}{metrics from \cite{rottmann18} } & \multicolumn{2}{c||}{entropy baseline}          \\
\cline{1-7}                                                                  
                  & train                &  val                     & train                &  val                           & train                &  val                         \\
\cline{1-7}                                                                                                                                                                          
ACC               & $79.88\%(\pm0.21\%)$ & $\boldsymbol{79.58\%(\pm0.15\%)}$     & $77.37\%(\pm0.28\%)$ & $77.24\%(\pm0.29\%)$           & $70.16\%(\pm0.22\%)$ & $70.16\%(\pm0.22\%)$           \\
AUROC             & $87.61\%(\pm0.16\%)$ & $\boldsymbol{87.38\%(\pm0.16\%)}$     & $85.32\%(\pm0.22\%)$ & $85.18\%(\pm0.20\%)$           & $77.73\%(\pm0.21\%)$ & $77.69\%(\pm0.21\%)$           \\
\cline{1-7}                                                                                                                                                                                      
\multicolumn{7}{||c||}{Meta Regression $IoU_{adj}$} \\                            
\cline{1-7}                                                                  
$\sigma$          & $0.135(\pm0.001)$    & $\boldsymbol{0.135(\pm0.001)}$        & $0.144(\pm0.001)$    & $0.144(\pm0.001)$              & $0.213(\pm0.001)$    & $0.213(\pm0.001)$              \\
$R^2$             & $81.72\%(\pm0.22\%)$ & $\boldsymbol{81.71\%(\pm0.20\%)}$     & $79.00\%(\pm0.20\%)$ & $79.08\%(\pm0.20\%)$           & $54.30\%(\pm0.40\%)$ & $54.43\%(\pm0.28\%)$           \\
\cline{1-7}                                                                         
\end{tabular}
}}
\caption{Results for all for meta classification and regression for three different sets of metrics. (Left block): all metrics including variance metrics, (center block): all metrics but without variances, (right-hand block): set of metrics used in \cite{rottmann18}, i.e., not marked with a ${}^*$ in \cref{tab:corr_coeff}. The best results for the validation set are highlighted.} 
\label{tab:results_all_metrics}
\end{table*}

%\vspace{-2.5ex}
%\newpage
%\vspace{1ex}
%\noindent
%\textbf{Correlation of the segment-wise metrics with the $\boldsymbol{\sIoU}$.}
\paragraph{Correlation of segment-wise metrics with the $\boldsymbol{\sIoU}$.}
\Cref{tab:corr_coeff} contains the Pearson correlation coefficients for all segment-wise metrics for all 16 available image crops constructed in \cref{sec:metrics}. We observe strong correlations for the measures $\bar D$ and $\bar D_\intr$ where $D \in \{ \mean M,  \mean V, \mean  E \}$ and for the relative size measures $\tilde S$ and $\tilde S_\intr$. All other size measures as well as $\mean D_\bdr$ for $D \in \{ M, V, E \} $ also show increased correlation coefficients. The variances and the Kullback-Leibler measures seem to play a minor role, however they might contribute additional information for a model that predicts the $\sIoU$.

%A variable selection via a lasso regression reveals that a strong subset with a $10$ metrics is \commentMR{hier ein lasso subset nennen}.

%\vspace{-2.5ex}
%\vspace{1ex} \noindent
%\textbf{Metric selection for meta classification and meta regression.}
\paragraph{Metric selection for meta classification and meta regression.}
In \cref{tab:subset_comparison} we compare different subsets of metrics. For the tasks of meta classification, we do so in terms of meta classification accuracy ($\sIoU=0$ vs.\ $\sIoU>0$) and in terms of the area under curve corresponding to the receiver operator characteristic curve (AUROC, see~\cite{DBLP:conf/icml/DavisG06}). The receiver operator characteristic curve is obtained by varying the decision threshold of the classification output for deciding whether $\sIoU=0$ or $\sIoU>0$. For the task of meta regression we state resulting standard deviations $\sigma$ of the linear regression fit's residual as well as $R^2$ values. We observe that the probability margin heat map yields the most predictive set of metrics, closely followed by the variation ratio. Altogether all heat maps yield fairly similar results and also the segment sizes yield a strong predictive set. The mean class probabilities $P$ by itself are not predictive enough, at least for linear and logistic regression models as being used here. In all cases we observe a significant performance increase when incorporating the variance based heat maps, also the geometric center yields valuable extra information. When using all metrics together, another significant increase in all performance measures can be observed. Noteworthily, we obtain AUROC values of up to $87.38\%$ for meta classification and $R^2$ values of up to $81.71\%$ for meta regression which demonstrates the predictive power of our metrics. When omitting the variance based metrics, the performance can not be maintained entirely, i.e., we observe a slight decrease of $0.28$ to $0.35$ pp in all accuracy measures. A visual demonstration of the meta regression performance can be found in \cref{fig:illus1}.

In order to further analyze the different subsets of metrics, we perform a greedy heuristic. We start with an empty set of metrics and add iteratively a single metric that improves meta prediction performance maximally. We perform this test twice, once for meta classification accuracy and once for meta regression $R^2$. \Cref{fig:var_select} depicts both performance measures as functions of the number of metrics. In both cases the curves stagnate quite quickly, indicating that a small set of metrics might be sufficient for a good model. This is confirmed by the results stated in \cref{tab:greedy_var_select}.
For the meta regression four of the first six metrics are variants of the probability margin. Combined with the geometric center $k_2$ and the relative segment size $\tilde{S}$, this set obtains an $R^2$ of $80.23\%$. Adding the rest of the metrics to this set only results in an increase of $1.48$ pp to the final $R^2$ of $81.71\%$. For the meta classification we start with $\tilde{S}$ at $77.25\%$ classification accuracy which is only $2.33$ pp below the accuracy for all metrics. Six out of the first ten added metrics are class probabilities and already after the seventh metric we obtain a classification accuracy of $79.18\%$. In both cases, for meta classification and regression, a small subset of metrics can be determined such that the corresponding performance is close to the performance for the full set of metrics. Also in both cases the variation ratio heat map $V_z$ is not required.

%\vspace{-2.5ex}
%\vspace{1ex} \noindent
%\textbf{Comparison with baseline approaches and others.}
\paragraph{Comparison with baseline approaches and others.}
In \cref{tab:results_all_metrics} we compare our results for all metrics with the set metrics introduced in \cite{rottmann18} (cf.~\cref{tab:corr_coeff}) and an entropy baseline where only a single entropy metric $\mean \bar E$ is employed. We do so as the entropy is a very commonly used uncertainty measure.
%We complement the results for the full set of metrics with another set where we leave out all variance based metrics and a third set that is introduced in \cite{rottmann18}.
In terms of AUROC we obtain an improvement of $2.20$ pp and in terms of $R^2$ of $2.63$ pp. When comparing the full set of metrics with the entropy baseline we obtain very pronounced gaps, $9.69$ pp in AUROC and $27.28$ pp in $R^2$.
In all three cases training and validation accuracies are tight, i.e., we do not observe any overfitting issues.

%In terms of AUROC we obtain an improvement over the entropy baseline of up to $10$ pp and in terms of $R^2$ up to $30\%$.
%When employing all metrics, we observe a slight increase of $0.28\%$ compared to the metrics set without variances.
%The improvements over the results for the metrics from \cite{rottmann18} are significant in all performance measures as well. In 

\begin{table}[t]
\scalebox{0.73}{{\setlength{\extrarowheight}{0.75pt}%
\begin{tabular}{||l||c|c||c||}
\cline{1-4}
\multicolumn{4}{||c||}{Classification $IoU_{adj}=0,>0$} \\
\cline{1-4}
                             & \multicolumn{2}{c||}{neural networks}             & \multicolumn{1}{c||}{linear models}     \\
\cline{1-4}                                                                       
                             & train                &  val                          &     val                   \\
\cline{1-4}                                                                                                                        
ACC                          & $83.22\%(\pm0.15\%)$ & $\boldsymbol{81.93\%(\pm0.22\%)}$          & $79.58\%(\pm0.15\%)$   \\
AUROC                        & $91.00\%(\pm0.11\%)$ & $\boldsymbol{89.89\%(\pm0.07\%)}$          & $87.38\%(\pm0.16\%)$   \\
\cline{1-4}                                                                                                                        
\multicolumn{4}{||c||}{Regression $IoU_{adj}$} \\                                 
\cline{1-4}                                                                       
$\sigma$                     & $0.120(\pm0.000)$    & $\boldsymbol{0.123(\pm0.001)}$             & $0.135(\pm0.001)$      \\
$R^2$                        & $85.48\%(\pm0.07\%)$ & $\boldsymbol{84.77\%(\pm0.30\%)}$          & $81.71\%(\pm0.20\%)$   \\
\cline{1-4}                                                                         
\end{tabular}
}}
\caption{Results obtained from a neural network used for meta classification and meta regression with all metrics. For simpler comparison we state the validation accuracies for linear models. The best results for the validation set are highlighted. \label{tab:results_nn_and_linear_model}
}%
\end{table}

%\vspace{-2.5ex}
%\vspace{1ex} \noindent
%\textbf{Meta classification and regression with neural networks.}
\paragraph{Meta classification and regression with neural networks.}
We repeat the tests from \cref{tab:results_all_metrics} for all metrics, however this time we use neural networks for meta classification and regression. Our neural networks are equipped with two hidden layers containing $61$ neurons each and we employ $\ell_2$ regularization with $\lambda=0.005$, results are stated in \cref{tab:results_nn_and_linear_model}. The difference between training and validation accuracies indicates that the neural network is slightly overfitting. When deploying neural networks instead of linear models, the validation accuracy increases by $2.35$ pp and the validation AUROC by $2.51$ pp. For the meta regression, the standard deviation $\sigma$ is reduced by $0.012$ and the $R^2$ value is increased significantly by $3.06$ pp. Note that, the results for $\sigma$ may lack interpretability when using a neural network, just as the whole model trades transparency for performance.

\section{Conclusion and Outlook}

In this paper we extend the approach presented in \cite{rottmann18}. Firstly, we introduce an approach that generates a batch of nested image crops that are presented to the segmentation network and yield a batch of probability distributions. The aggregated probabilities show improved $\mIoU$ values, especially with respect to the far range section in the center of the input image.  Secondly, we add segment-wise metrics constructed from variation ratio, Kullback-Leibler divergence, geometric center and crop variance based metrics. Thirdly, for the meta classification and meta regression, we replace the linear model with neural networks. All three aspects contribute to a significant improvement over the approach presented in \cite{rottmann18}. More precisely, we obtain an increase in meta classification accuracy of $4.69$ pp and an increase of AUROC of $4.80$ pp. The $R^2$ for meta regression is increased by $5.69$ pp. 
Currently we are working on time-dynamic meta classification and regression approaches which make predictions from time series of metrics. As we only presented an approach for false positive detection we also plan to combine this with approaches for false negative detection, see e.g.~\cite{chan19_1}. Combining these approaches might eventually result in improved segmentation performance, at least with respect to certain classes. The source code of our method is publicly available at \url{https://github.com/mrottmann/MetaSeg/tree/nested_metaseg}.

%\vspace{1ex} \noindent
%\textbf{Acknowledgements.}
\paragraph{Acknowledgements.}
We thank Hanno Gottschalk, Peter Schlicht and Fabian H\"uger for discussion and useful advice.

{\small
\bibliographystyle{ieee}
\bibliography{biblio}
}

\end{document}